\documentclass[a4paper,fleqn]{cas-dc}

\usepackage[linesnumbered,ruled,vlined]{algorithm2e}

\usepackage{svg}
\usepackage{pdfpages}
\usepackage[numbers]{natbib}

\def\tsc#1{\csdef{#1}{\textsc{\lowercase{#1}}\xspace}}
\tsc{WGM}
\tsc{QE}
\tsc{EP}
\tsc{PMS}
\tsc{BEC}
\tsc{DE}

\begin{document}
\let\WriteBookmarks\relax
\def\floatpagepagefraction{1}
\def\textpagefraction{.001}
\shorttitle{Imbalanced Image Classification with Complement Cross Entropy}
\shortauthors{Y. Kim et~al.}

\title [mode = title]{Imbalanced Image Classification with Complement Cross Entropy}

\author[1]{Yechan Kim}
[orcid=0000-0002-2438-3590]
\ead{yechankim@gm.gist.ac.kr}

\address[1]{Gwangju Institute of Science and Technology (GIST), Republic of Korea}


\author[1]{Younkwan Lee}
[orcid=0000-0001-5622-3383]
\ead{brightyoun@gist.ac.kr}

\author[1]{Moongu Jeon} 
[orcid=0000-0002-2775-7789]
\ead{mgjeon@gist.ac.kr}
\cormark[1]

\cortext[cor1]{Corresponding author.}

\begin{abstract}
    Recently, deep learning models have achieved great success in computer vision applications, relying on large-scale class-balanced datasets. 
    However, imbalanced class distributions still limit the wide applicability of these models due to degradation in performance. 
    To solve this problem, in this paper, we concentrate on the study of cross entropy which mostly ignores output scores on incorrect classes. 
    This work discovers that neutralizing predicted probabilities on incorrect classes improves the prediction accuracy for imbalanced image classification. 
    This paper proposes a simple but effective loss named \textit{complement cross entropy} based on this finding.
    The proposed loss makes the ground truth class overwhelm the other classes in terms of softmax probability, by neutralizing probabilities of incorrect classes, without additional training procedures. 
    Along with it, this loss facilitates the models to learn key information especially from samples on minority classes. 
    It ensures more accurate and robust classification results on imbalanced distributions. 
    Extensive experiments on imbalanced datasets demonstrate the effectiveness of the proposed method.
\end{abstract}



\begin{keywords}
Loss function\sep
Deep learning\sep
Class imbalance\sep
Image classification\sep
Complement cross entropy
\end{keywords}

\maketitle

\section{Introduction}
    In recent years, computer vision algorithms led by deep neural networks (DNNs) have achieved remarkable success in many tasks such as image classification \cite{krizhevsky2012imagenet, simonyan2014very, szegedy2015going, he2016deep}, sequence generation \cite{sutskever2014sequence, bahdanau2014neural} and text recognition \cite{cheng2018aon, lee2019snider, lee2020self}.
    Such a widespread adoption is attributable to (i) the existence of large-scale datasets with a vast number of annotations and (ii) the use of cross entropy as a standard training objective.
    However, various emerging datasets typically exhibit extremely imbalanced class distributions, which largely limit the capability of the DNN model in terms of generalization.
    Although such imbalanced distributions in the existing real-world data is obviously a crucial challenge, not much research has been conducted yet.

    To solve this issue, one common strategy is to resample the dataset, \emph{e.g.}, oversampling on minority classes \cite{chawla2002smote, he2008adasyn, piras2012synthetic,  castellanos2018oversampling, kim2019valid, sadhukhan2019reverse},
    undersampling on majority classes \cite{drummond2003c4, yen2009cluster, fan2016one, koziarski2020radial, liu2020dealing}, and a hybrid of both \cite{batista2004study, zeng2016effective, tang2017gir,zhu2020inspector}. 
    Another approach is to employ cost sensitive learning, \emph{e.g.}, reweighting sample-wise loss by inverse class frequency and penalizing hard-classified samples (typically, minority classes) by assigning relatively higher loss \cite{lin2017focal, shafieezadeh2019regularization, ryou2019anchor, wang2020imbalance}. However, these approaches typically neglect the fact that samples on minority classes may have noise or false annotations. It means that training criterion which largely focuses on the minority classes rather than majority classes might cause poor generalization of the model \cite{ren2018learning}.

    \begin{figure}[ht]
    \centering
    \includegraphics[width=8.3cm]{ 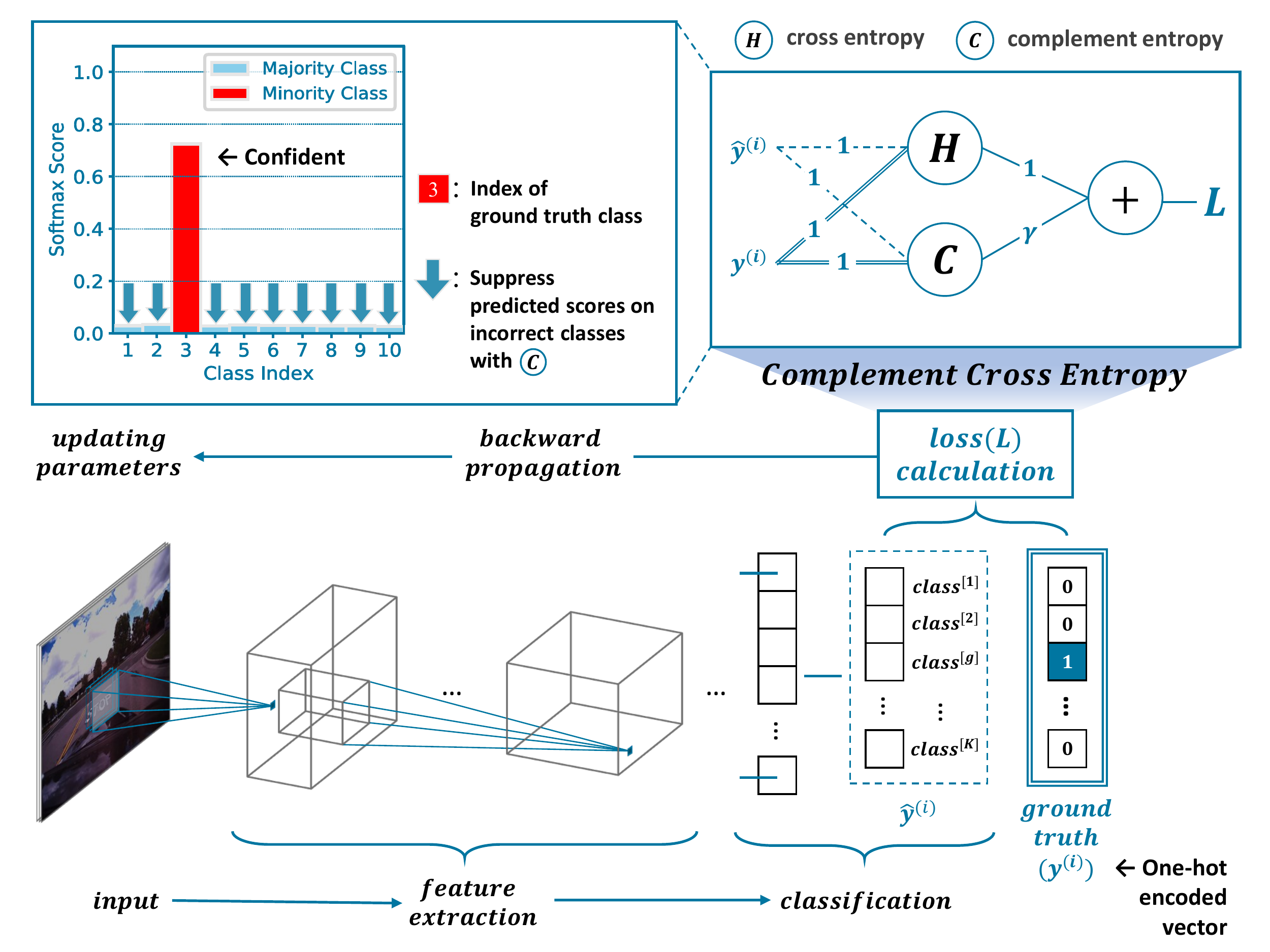}
    \caption{Overview of training a classification model with the proposed loss function, named \textit{Complement Cross Entropy} (CCE). CCE attempts to suppress predicted probabilities on incorrect classes. This encourages the model to extract the essential features particularly in imbalanced datasets. It ensures robust classification results because it provides better training chances for samples on minority classes.}
    \label{fig1}
    \vspace{-0.5cm}
    \end{figure}

    In order to address this problem, this work revisited the cross entropy and observed many degradation problems in imbalanced datasets. 
    Cross entropy (or KL divergence \cite{kullback1951information}) is one of the most well-known loss functions. 
    For instance, softmax cross entropy is generally used for classification model optimization. 
    Besides, sigmoid cross entropy is widely adopted in the object detection task. 
    This work mainly focuses on softmax cross entropy.
    The softmax cross entropy, $H(\mathbf{y}, \hat{\mathbf{y}})$ is defined as:
    \begin{equation}\label{eq1}
    \begin{aligned}
        H(\mathbf{y}, \hat{\mathbf{y}}) 
        = & -\frac{1}{N}\sum_{i=1}^{N}\sum_{j=1}^{K} {\mathbf{y}}^{(i)}_{[j]} \log {\hat{\mathbf{y}}}^{(i)}_{[j]}, \\
    \end{aligned}
    \end{equation}
    where $N$ is the number of examples, $K$ is the number of categories, $\mathbf{y}^{(i)}$ is the true distribution, and $\hat{\mathbf{y}}^{(i)}$ is the softmax multinomial prediction distribution.
    Minimizing this softmax loss leads to maximizing the predicted probability of the ground truth class, which would be a good property for model optimization. 
    However, this loss might be insufficient to explicitly minimize the predicted probabilities of incorrect classes, particularly for minority classes.
    The above-mentioned analysis is based on the fact that all softmax probabilities on incorrect classes are ignored in Eq. (1), because ${\mathbf{y}}^{(i)}_{[j \ne g]}$ (where $g$ denotes the ground truth index) is always zero when $\mathbf{y}^{(i)}$ is a one-hot represented vector. 
    It means that $\hat{\mathbf{y}}^{(i)}_{[j \ne g]}$ in Eq. (1) is totally ignored, so inaccurately predicted probabilities may produce a cumulative error, particularly in the class-imbalanced distribution. 

    To avoid such errors, we introduce a novel loss named \textbf{complement cross entropy (CCE)} to tackle such performance degradation problem on imbalanced dataset. 
    It is motivated by \textit{complement objective training} \cite{chen2019complement}, where the core idea is evenly suppressing softmax probabilities on incorrect classes during training. 
    The proposed loss does not require additional augmentation of samples or upscaling loss scales for the minority classes. 
    Besides, this does not require hard hyperparameter tuning efforts, which would be a good property to apply to an unknown dataset.
    Instead, the proposed method utilizes information on incorrect classes to train a robust classification model for imbalanced class distribution. 
    Therefore, this work argues that this strategy provides better learning chances particularly for samples on minority classes because it encourages the correct class (including minority one) to overwhelm its softmax score across all the other ``incorrect'' classes.

    The main contributions of this work are summarized as follows. 
    (i) This paper presents a new perspective for imbalanced classification: neutralizing the probability distribution on incorrect classes enables more robust classification on class-imbalanced scenarios. Based on this view, this work proposes a new loss function that efficiently reduces the risk of misprediction, particularly in minority classes. 
    (ii) This work experimentally demonstrates the effectiveness of the proposed method for classification on imbalanced datasets. 
    The proposed loss boosts the classification accuracy and also provides a faster convergence speed in some cases. 

    The rest of this paper is organized as follows: 
    In Section 2, this paper first introduces a key concept of complement entropy and presents the proposed loss function. 
    Section 3 provides an overview of datasets and implementation details for experiments, and then presents the experimental results of the proposed method. 
    Finally, conclusion and future work are described in Section 4.

\section{Method}
    This section first provides a brief concept of complement entropy and then presents the proposed method for imbalanced image classification.

\subsection{Complement Entropy}
    Complement entropy is designed to encourage models to learn enhanced representations by assisting the primary training objective, cross entropy. 
    It is calculated as a mean of Shannon's entropies on incorrect classes of the whole examples.
    The complement entropy, $C(\hat{\mathbf{y}})$ is formulated as:
    \begin{equation}\label{eq2}
    \begin{aligned}
        C(\hat{\mathbf{y}}) =
        -\frac{1}{N}\sum_{i=1}^{N}\sum_{j = 1, j \ne g}^{K} 
        \frac{\hat{\mathbf{y}}^{(i)}_{[j]}}{1-\hat{\mathbf{y}}^{(i)}_{[g]}}  
        \log \frac{\hat{\mathbf{y}}^{(i)}_{[j]}}{1-\hat{\mathbf{y}}^{(i)}_{[g]}} ,
    \end{aligned}
    \end{equation}
    where $g$ represents the ground truth index.
    In Eq. (2), the inverse of ($1-\hat{\mathbf{y}}^{(i)}_{[g]}$) normalizes $\hat{\mathbf{y}}^{(i)}_{[j]}$ to make $C(\hat{\mathbf{y}})$ imply the information underlying probability distribution on just incorrect classes. 
    The purpose of this entropy is to encourage the predicted probability of the ground truth class ($\hat{\mathbf{y}}^{(i)}_{[g]}$) to be larger among the other incorrect classes.
    This means that the more the model neutralizes the distribution of predicted probabilities for the incorrect classes during learning, the more confident the prediction for the correct class ($\hat{\mathbf{y}}^{(i)}_{[g]}$) becomes (see Fig. \ref{fig1}). 
    To this end, an optimizer should maximizes complement entropy in Eq. (2), since the Shannon's entropy becomes maximized when the probability distribution is uniform (or flattened).
    %


    This work is motivated by a concept of complement entropy. 
    With adopting this concept, the predicted probability on the ground truth class is less vulnerable to the probabilities on the other incorrect classes. 
    When the minority class is the ground truth class, this mechanism allows the model to find the better hidden pattern from samples of the minority class because it prevents the minority classes from being threatened by the rest incorrect classes (including majority ones) during the training process.

    Before going further, this paper defines balanced complement entropy. This entropy, $C'(\hat{\mathbf{y}})$ is designed to match the scale between cross entropy and complement entropy and formulated as:
    \begin{equation}\label{eq3}
    \begin{aligned}
        C'(\hat{\mathbf{y}}) =
        \frac{1}{K-1} C(\hat{\mathbf{y}}),
    \end{aligned}
    \end{equation}
    where $\frac{1}{K-1}$ is the balancing factor.

    \begin{algorithm}
    \caption{Training with a bi-objective concept: cross entropy and complement entropy}
    
    \SetKwInOut{Input}{input}
    \SetKwInOut{Output}{output}
    \Input{Training dataset, {$\mathbf{D}=\{(\mathbf{X}_{i}, \mathbf{y}_{i})\}^{N}_{i=1} $}}
    \Output{Model parameters, \{$\mathbf{\theta}_{1}$, $\cdots$, $\mathbf{\theta}_{n_{layers}}$\}}
     initialization\;
      \For{$t \leftarrow 1$ $\emph{\textbf{to}}$ $n_{train\_steps}$}{
            $\mathbf{X}, \mathbf{y}$ $\leftarrow$ mini\_batch($\mathbf{D}, t$)\;
            $\hat{\mathbf{y}}$ $\leftarrow$ model($\mathbf{X}, \mathbf{y}$)\;
            primary\_optimizer.step($ H(\mathbf{y}, \hat{\mathbf{y}}) $)\;
            secondary\_optimizer.step($ C'(\hat{\mathbf{y}}) $)\;
     }
    \end{algorithm}

    Algorithm 1 describes how to train neural networks with cross entropy (primary training objective) and complement entropy (complement training objective). 
    At each iteration in training, cross entropy, $H(\mathbf{y}, \hat{\mathbf{y}})$ is first used to update the model parameters; (balanced) complement entropy, $C'(\hat{\mathbf{y}})$ is then needed to update the parameters again. Extensive experiments have already been conducted by Chen \textit{et al.} and they demonstrate the effectiveness of complementing cross entropy with complement entropy for stable training \cite{chen2019complement}. Despite its efficacy, it has one crucial limitation: it induces a training time approximately two times longer because it requires twice back-propagation per each iteration in this training mechanism. On the other hand, this paper proposes a single training loss function that efficiently performs like the above training concept. It allows the model optimizer to back-propagate only once rather than twice at each iteration.

\subsection{Complement Cross Entropy (CCE)}
    In contrast to the algorithm 1, this work replaces the training process by combining cross entropy and complement entropy with a single entropy (see line 5 in algorithm 2). 
    Algorithm 2 depicts a training procedure with the proposed loss. 
    Experiments in this work show that the proposed method requires a shorter training time: Algorithm 2 performs about 1.7 times faster than algorithm 1. 
    To balance cross entropy and complement entropy, this work adds $\gamma$ to the complement entropy as:
    \begin{equation}\label{eq4}
    \begin{aligned}
        \Tilde{C}(\hat{\mathbf{y}}) =
        \frac{\mathbf{\gamma}}{K-1} C(\hat{\mathbf{y}}),
    \end{aligned}
    \end{equation}
    where the modulating factor, $\gamma$ should be tuned to decide the amount that complements the cross entropy, \textit{e.g.}, $\gamma=-1$ ($\gamma<0$). The proposed loss, named complement cross entropy (CCE), is defined as:
    \begin{equation}\label{eq5}
    \begin{aligned}
        H(\mathbf{y}, \hat{\mathbf{y}}) + {\Tilde{C}}(\hat{\mathbf{y}}).
    \end{aligned}
    \end{equation}

    \vspace{-0.5cm}
    \begin{algorithm}
    \SetKwInOut{Input}{input}
    \SetKwInOut{Output}{output}
    \Input{Training dataset, {$\mathbf{D}=\{(\mathbf{X}_{i}, \mathbf{y}_{i})\}^{N}_{i=1} $}}
    \Output{Model parameters, \{$\mathbf{\theta}_{1}$, $\cdots$, $\mathbf{\theta}_{n_{layers}}$\}}
    
     initialization\;
      \For{$t \leftarrow 1$ $\emph{\textbf{to}}$ $n_{train\_steps}$}{
            $\mathbf{X}, \mathbf{y}$ $\leftarrow$ mini\_batch($\mathbf{D}, t$)\;
            $\hat{\mathbf{y}}$ $\leftarrow$ model($\mathbf{X}, \mathbf{y}$)\;
            $\mathbf{final\_loss}$ $\leftarrow$ $H(\mathbf{y}, \hat{\mathbf{y}})$ + ${\Tilde{C}}(\hat{\mathbf{y}})$\;
            optimizer.step($\mathbf{final\_loss}$)\;
     }
     \caption{Training with CCE loss (Proposed)}
     \label{alg:2}
    \end{algorithm}

\vspace{-0.5cm}
\section{Experiments}
    \begin{figure*}[ht]
    \centering\includegraphics[width=17cm]{ 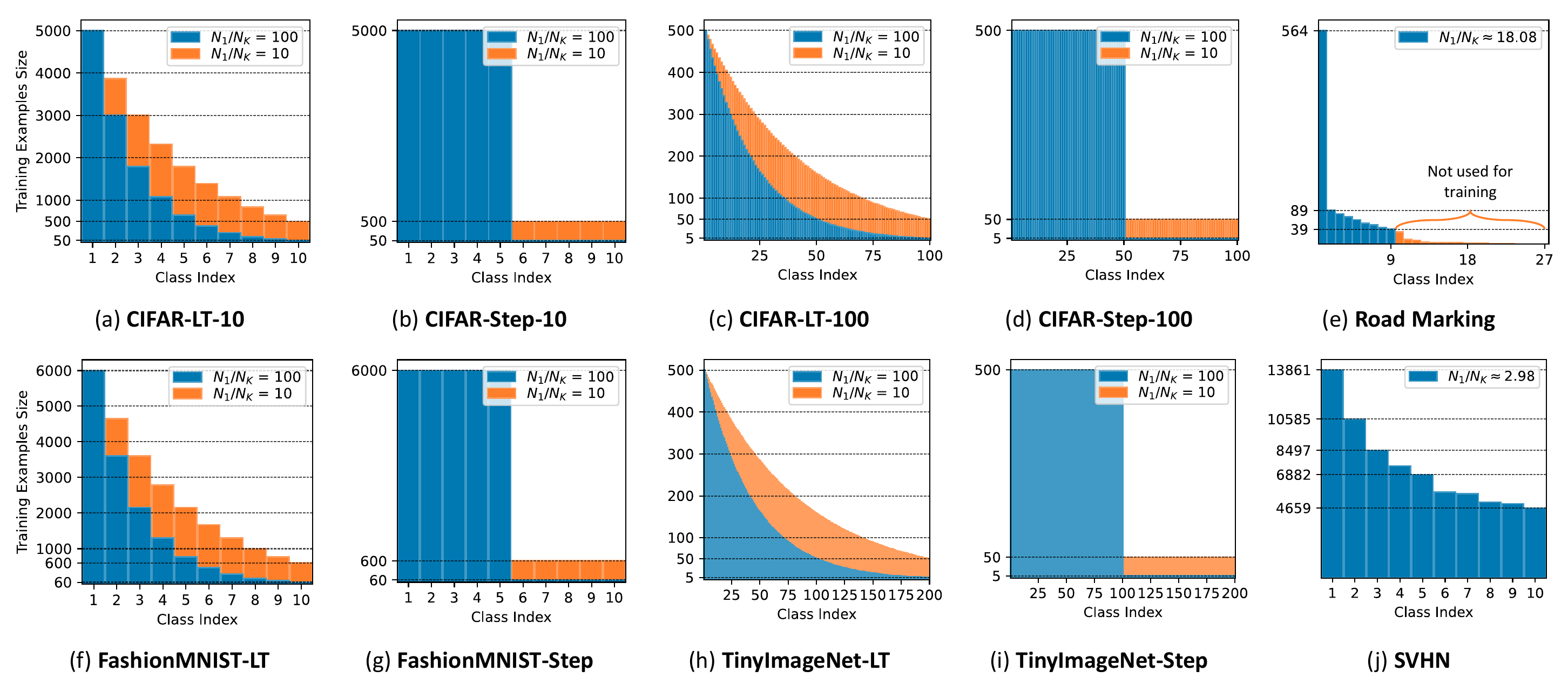}
    \caption{Class distributions of training examples on (i) imbalanced variants of CIFAR-10, CIFAR-100, Fashion MNIST and Tiny ImageNet, and (ii) Road Marking and SVHN. Note that ``LT'' and ``Step'' represent a long-tailed distribution and a step distribution, respectively, and each class index is sorted in descending order by the sample size. 
    }
    \label{fig2}
    \end{figure*}

    This section first briefly overviews the experimental setup and implementation details and then presents experimental results per each imbalanced image dataset.
    
    \subsection{Imbalance Setting}
    In order to evaluate the proposed method on various levels of imbalance, this work constructs synthetically imbalanced variants of well-refined datasets, \textit{e.g.}, CIFAR, Fashion MNIST and Tiny ImageNet, if underlying class distributions are balanced.
    Basically, imbalanced variants are constructed by randomly removing examples per each class.
    More specifically, this work considers two types of imbalance in exactly a same way as \cite{cao2019learning}: (a) \textit{long-tailed distribution}: making the train sets to follow an exponential decay distribution in sample sizes per each class; (b) \textit{step distribution}:  making the half of classes to own the same amount of huge training examples, while the rest of classes own the same size of less training examples, consequently leading to a step-shaped distribution (see Fig. \ref{fig2}). 
    For variants which are originally class-balanced, each test set still has a class-balanced distribution. 
    Note that $\frac{N_1}{N_K}$ represents imbalance ratio, where $N_i$ denotes training example size of $i^{th}$ class, $K$ is the maximum number of class index, and each class index is sorted in descending order by class-wise example size.

    \subsection{Evaluation Metric}
    \begin{table}[h]
    \footnotesize
    \caption{Confusion matrix for multi-class classification.}
    \centering
    \begin{tabular}{@{}cccccc@{}}
    \toprule
                                                 &         & \multicolumn{4}{c}{Predicted}     \\ \cmidrule(l){3-6} 
                                                 &         & Class $1$ & Class $2$ & $\cdots$ & Class $K$ \\ \midrule
    \multicolumn{1}{c|}{\multirow{4}{*}{Actual}} & \multicolumn{1}{c|}{Class $1$} & $N_{1, 1}$ & $N_{1, 2}$ & $\cdots$ & $N_{1, K}$ \\
    \multicolumn{1}{c|}{}                        & \multicolumn{1}{c|}{Class $2$} & $N_{2, 1}$ & $N_{2, 2}$ & $\cdots$ & $N_{2, K}$ \\
    \multicolumn{1}{c|}{}                        & \multicolumn{1}{c|}{$\vdots$}  & $\vdots$ & $\vdots$ & $\ddots$ & $\vdots$ \\
    \multicolumn{1}{c|}{}                        & \multicolumn{1}{c|}{Class $K$} & $N_{K, 1}$ & $N_{K, 1}$ & $\cdots$ & $N_{K, K}$ \\ \bottomrule
    \end{tabular}
    \label{metric1}
    \end{table}
    
    In order to evaluate the image classification performance, this work adopts balanced accuracy ($bACC$) \cite{brodersen2010balanced}. 
    Since $bACC$ is calculated as mean of correctly classified examples per each category, it avoids performance evaluation from being biased by majority classes on imbalanced datasets.
    In Table \ref{metric1}, $N_{i,i}$ ($1 \leq i \leq K$) is the number of examples correctly classified as class $i$; while $N_{i,j}$ ($1 \leq j \leq K$, $i \ne j$) is the number of examples incorrectly identified as class $j$. 
    $bACC$ is then formulated as:
    \begin{equation}\label{eq7}
    \begin{aligned}
        bACC = \frac{1}{K}\sum_{i=1}^{K}recall(i),
    \end{aligned}
    \end{equation}
    where $recall(i)$ is calculated as $N_{i,i} / \sum_{j=1}^{K}N_{i,j}$.
    

    \subsection{Experimental Setup}
    \textbf{Existing Methods.} This work compares the proposed method (CCE) with the following techniques: 
    (i) Empirical Risk Minimization (ERM): softmax cross entropy; 
    (ii) Complement Objective Training (COT): softmax cross entropy and complement entropy \cite{chen2019complement}; 
    (iii) Focal Loss (FL): softmax cross entropy with a modulating factor to concentrate on hard samples \cite{lin2017focal}.
    Note that this work selects the well-known loss functions that can be simply applied to DNN models without additional learning procedures or hyperparameter tuning efforts.
    
    \textbf{Training Details.} 
    Before training, this work applies zero padding, random cropping, and horizontal flipping to the training sets, not the test sets.
    Both training sets and test sets are normalized with mean and variance of the training ones.
    All models are trained for 200 epochs on each dataset but 100 epochs on Road Marking.
    Stochastic gradient descent (SGD) is adopted to optimize the models, where momentum of weight is 0.9, weight decay is 5e-4, and mini-batch size is 128.
    For training, the learning rate is initially set to 1e-1 and dropped by a factor of 0.5 at 60, 120, and 160 epochs in the same manner of \cite{pereyra2017regularizing} except for Road Marking.
    Besides, learning rate warm-up strategy \cite{he2019bag} is also used for first 5 epochs.
    Hyperparameter, $\gamma$ should be tuned for CCE: this work sets $\gamma$ to -1 over all experiments (see Subsection 3.5).
    The convolutional neural networks (CNNs) used for experiments are as follows: 
    ResNet, SqueezeNet, ResNeXt, DenseNet and EfficientNet \cite{he2016deep, iandola2016squeezenet, xie2017aggregated, huang2017densely, tan2019efficientnet}.

    \subsection{Experimental Results}

    This subsection describes each dataset used in the experiments and then presents the experimental result for each dataset.
    Note that this work observes that performance is improved by just replacing ERM to COT, which is the motivation of this work.
    Besides, the proposed loss outperforms the other existing methods in all experiments.
    
\begin{figure*}[ht]
    \centering\includegraphics[width=12.5cm]{ 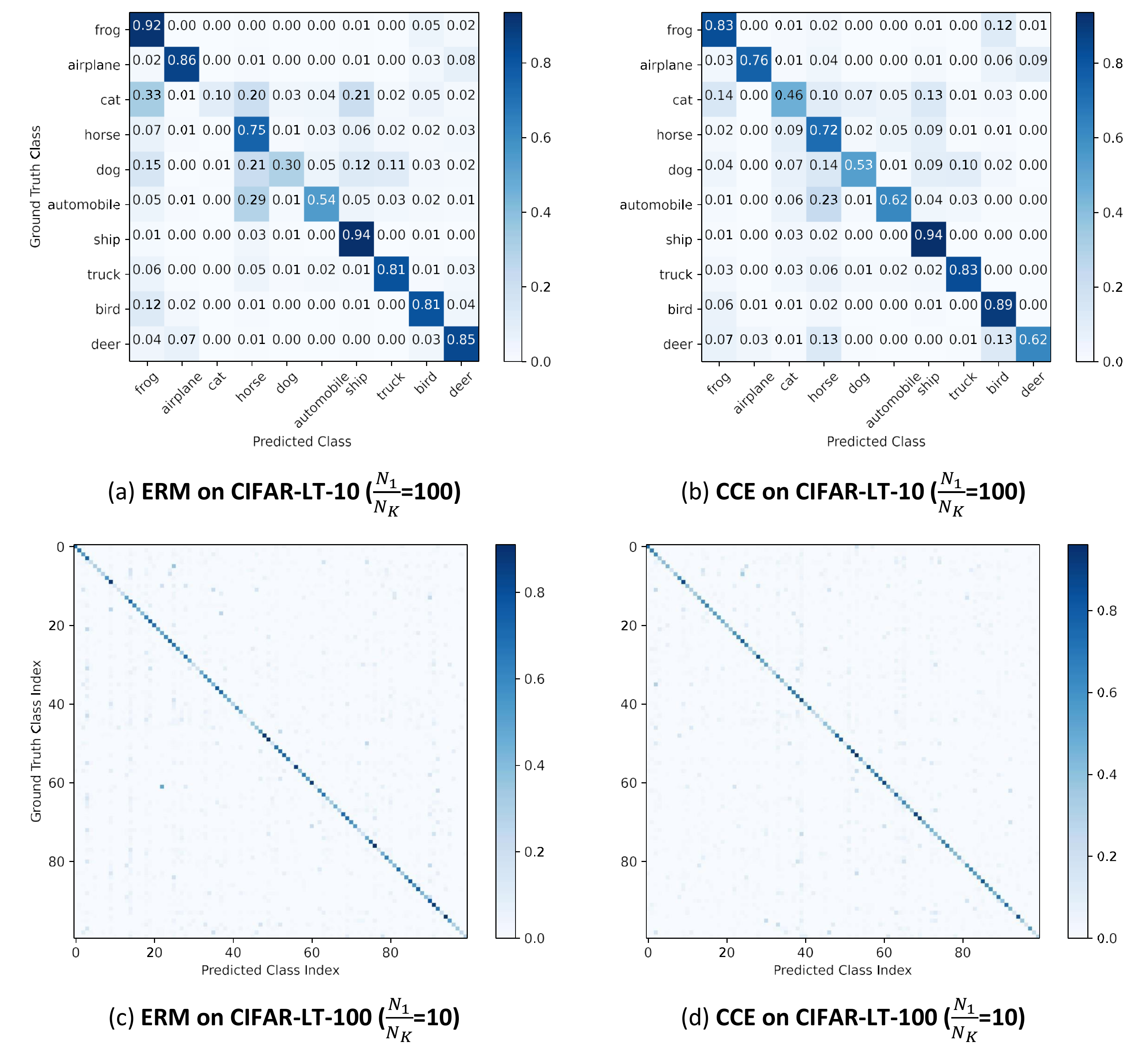}
    \caption{Confusion matrices of classification results on CIFAR-LT-10 and CIFAR-LT-100 (ResNet-34 is used as a backbone). 
    Remark that `cat,' `dog,' and `bird' are extremely minority classes in CIFAR-LT-10 ($\frac{N_1}{N_K}$=100).
    It can be seen that each accuracy value of diagonal term in matrix (a) becomes increased in the same positions of matrix (b), which means that the proposed loss helps samples on each minority class to be learned better by the models.
    Also, almost all diagonal components in matrix (b) and (d) are evenly darker than diagonal ones in matrix (a) and (c), respectively. 
    It indicates that CCE evenly improves the class-wise accuracy of prediction.
    }
    \label{fig-cifar}
    \vspace{-0.5cm}
    \end{figure*}

    \textbf{Imbalanced CIFAR-10 and CIFAR-100.}
    CIFAR-10 and CIFAR-100 \cite{krizhevsky2009learning} contain RGB images of real-world things (32$\times$32 pixels): 50,000 examples for training and 10,000 examples for testing. 
    Each number of classes on CIFAR-10 and CIFAR-100 is 10 and 100, respectively. 
    This work constructs the imbalanced variants as depicted in Fig. \ref{fig2} (a)-(d) because the original version of CIFAR is completely class-balanced.
    As shown in Table \ref{table-cifar-10}-\ref{table-cifar-100} and Figure \ref{fig-cifar}, the proposed loss outperforms the other methods in all experiments for imbalanced CIFAR with ResNet-34.

    \begin{table}[h]
    \footnotesize
    \centering
    \caption{Classification test accuracy (\%) on imbalanced variants of CIFAR-10 with ResNet-34.}
    \begin{tabular}{@{}lllll@{}}
    \toprule
                   & \multicolumn{4}{c}{CIFAR-10}\\ \cmidrule(l){2-5}
                   & \multicolumn{2}{c}{Long-tailed}                  & \multicolumn{2}{c}{Step}                         \\ \cmidrule(l){2-5} 
     Imbalance Ratio ($\frac{N_1}{N_K}$) & \multicolumn{1}{c}{10} & \multicolumn{1}{c}{100} & \multicolumn{1}{c}{10} & \multicolumn{1}{c}{100} \\ \midrule
     ERM & 87.21 & 68.80 & 85.98 & 67.21 \\
     FL  & 86.16 & 67.54 & 85.24 & 67.35 \\
     COT & 88.02 & 71.28 & 86.12 & 68.13 \\
     \textbf{CCE} & \textbf{88.37} & \textbf{71.98} & \textbf{86.85} & \textbf{68.58} \\
     \bottomrule
     \label{table-cifar-10}
     \end{tabular}
     \end{table}

    \begin{table}[h]
    \footnotesize
    \centering
    \caption{Classification test accuracy (\%) on imbalanced variants of CIFAR-100 with ResNet-34.}
    \begin{tabular}{@{}lllll@{}}
    \toprule
                      & \multicolumn{4}{c}{CIFAR-100}\\ \cmidrule(l){2-5}
                      & \multicolumn{2}{c}{Long-tailed}                  & \multicolumn{2}{c}{Step}                         \\ \cmidrule(l){2-5} 
    Imbalance Ratio ($\frac{N_1}{N_K}$) & \multicolumn{1}{c}{10} & \multicolumn{1}{c}{100} & \multicolumn{1}{c}{10} & \multicolumn{1}{c}{100} \\ \midrule
    ERM & 62.35 & 43.49 & 60.17 & 40.77 \\
    FL  & 63.10 & 43.68 & 60.94 & 40.85 \\
    COT & 62.59 & 43.94 & 60.42 & 40.74 \\
    \textbf{CCE} & \textbf{63.12}  & \textbf{44.21} & \textbf{61.02} & \textbf{40.85} \\
    \bottomrule
    \label{table-cifar-100}
    \end{tabular}
    \vspace{0.5cm}
    \end{table}

\begin{figure*}[ht]
    \centering\includegraphics[width=14.5cm]{ 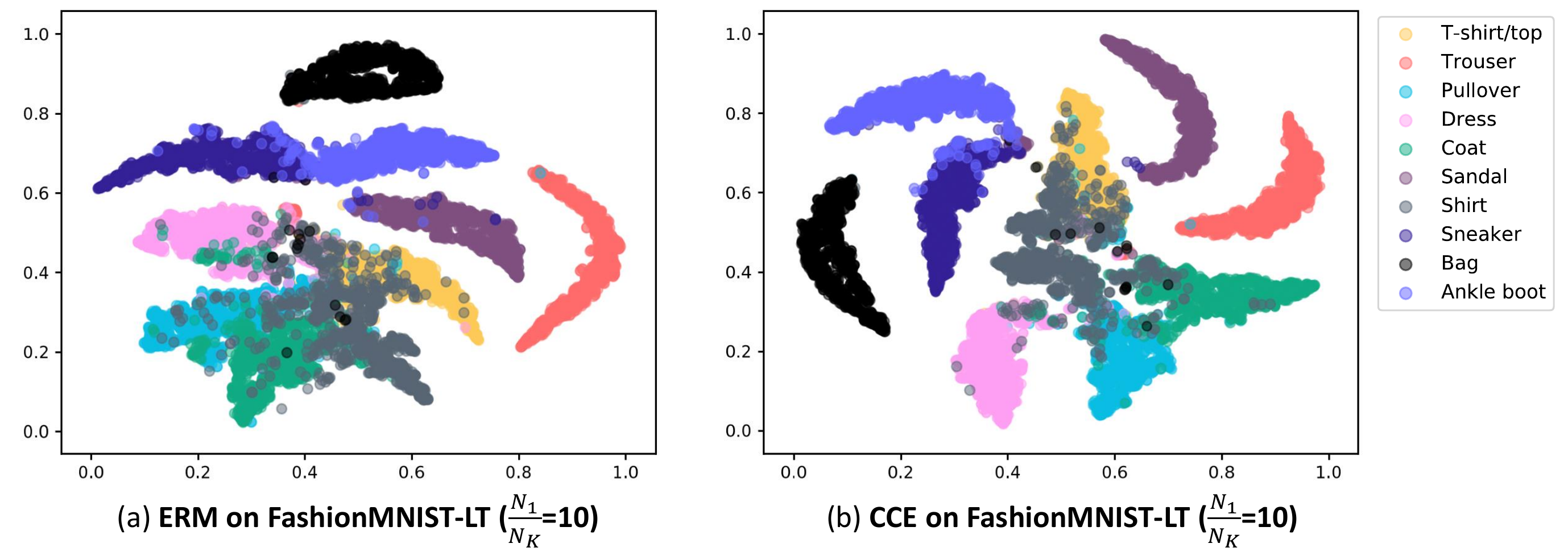}
    \caption{Feature embeddings of FashionMNIST-LT ($\frac{N_1}{N_K}$=10) with ResNet-34 trained by (a) ERM: softmax cross entropy and (b) CCE: complement cross entropy.
    This work extracts and flattens the last feature maps from the layer before the final classification layer.
    For visualization, this work then projects embedding representations into 2D vectors using t-SNE.
    Compared to (a), each cluster of `T-shirt/top,' `Pullover,' `Dress,' `Coat' and `Shirt' in (b) is farther apart in terms of inter-class distance.
    It means that CCE encourages the model to predict more accurately with more separable boundaries of categories.
    }
    \label{fig-tsne}
    \vspace{-0.5cm}
    \end{figure*}

    \textbf{Imbalanced Fashion MNIST.}
    Fashion MNIST \cite{xiao2017fashion} includes grayscale images of fashion products crawled from Zalando's website (28$\times$28 pixels): 60,000 examples for training and 10,000 examples for testing across 10 categories.
    The original version of Fashion MNIST is completely class-balanced, so this work constructs the imbalanced variants as depicted in Fig. \ref{fig2} (f)-(g). 
    Table \ref{table-fashion} and Fig. \ref{fig-tsne} show the efficacy of the proposed loss function in terms of prediction accuracy on imbalanced variants of Fashion-MNIST.
    
    %
    
    \begin{table}[h]
    \footnotesize
    \centering
    \caption{Classification test accuracy (\%) on imbalanced variants of Fashion MNIST with ResNet-34.}
    \begin{tabular}{@{}lllll@{}}
    \toprule
                      & \multicolumn{4}{c}{Fashion MNIST}\\ \cmidrule(l){2-5}
                      & \multicolumn{2}{c}{Long-tailed}                  & \multicolumn{2}{c}{Step}                         \\ \cmidrule(l){2-5} 
    Imbalance Ratio ($\frac{N_1}{N_K}$) & \multicolumn{1}{c}{10} & \multicolumn{1}{c}{100} & \multicolumn{1}{c}{10} & \multicolumn{1}{c}{100} \\ \midrule
    ERM & 91.45 & 87.98 & 91.08 & 85.54 \\
    FL  & 91.64 & 87.84 & 90.99 & 85.61 \\
    COT & 92.13 & 88.32 & 91.33 & 85.85 \\
    \textbf{CCE} &  \textbf{92.40} & \textbf{88.97} & \textbf{91.48} & \textbf{86.10} \\
    \bottomrule
    \label{table-fashion}
    \end{tabular}
    \end{table}
    
    \textbf{Imbalanced Tiny ImageNet.} 
    Tiny ImageNet contains colored real-world things (64$\times$64 pixels), which is a cropped version of ImageNet \cite{deng2009imagenet}: 100,000 examples for training and 10,000 examples for testing across 200 classes.
    The original version of Tiny ImageNet is completely class-balanced, so this work constructs the imbalanced variants as depicted in Fig. \ref{fig2} (h)-(i). 
    During testing, each image is centrally cropped in a 56$\times$56 size. 
    As indicated in Table \ref{table-tiny}, the proposed loss function is superior to the other existing methods for all experiments on imbalanced Tiny ImageNet.

    \begin{table}[h]
    \footnotesize
    \centering
    \caption{Classification test accuracy (\%) on imbalanced variants of Tiny ImageNet with ResNet-34.}
    \begin{tabular}{@{}lllll@{}}
    \toprule
                      & \multicolumn{4}{c}{Tiny ImageNet}\\ \cmidrule(l){2-5}
                      & \multicolumn{2}{c}{Long-tailed}                  & \multicolumn{2}{c}{Step}                         \\ \cmidrule(l){2-5} 
    Imbalance Ratio ($\frac{N_1}{N_K}$) & \multicolumn{1}{c}{10} & \multicolumn{1}{c}{100} & \multicolumn{1}{c}{10} & \multicolumn{1}{c}{100} \\ \midrule
    ERM & 49.41 & 32.80 & 47.68 & 35.04 \\
    FL  & 49.68 & 32.78 & 47.05 & 35.17 \\
    COT & 49.85 & 33.80 & 47.97 & \textbf{35.91} \\
    \textbf{CCE} &  \textbf{50.01} & \textbf{33.86} & \textbf{48.04} & 35.43 \\
    \bottomrule
    \label{table-tiny}
    \end{tabular}
    \vspace{-0.5cm}
    \end{table}

\begin{figure}[htbp]
    \centering\includegraphics[width=5.5cm]{ 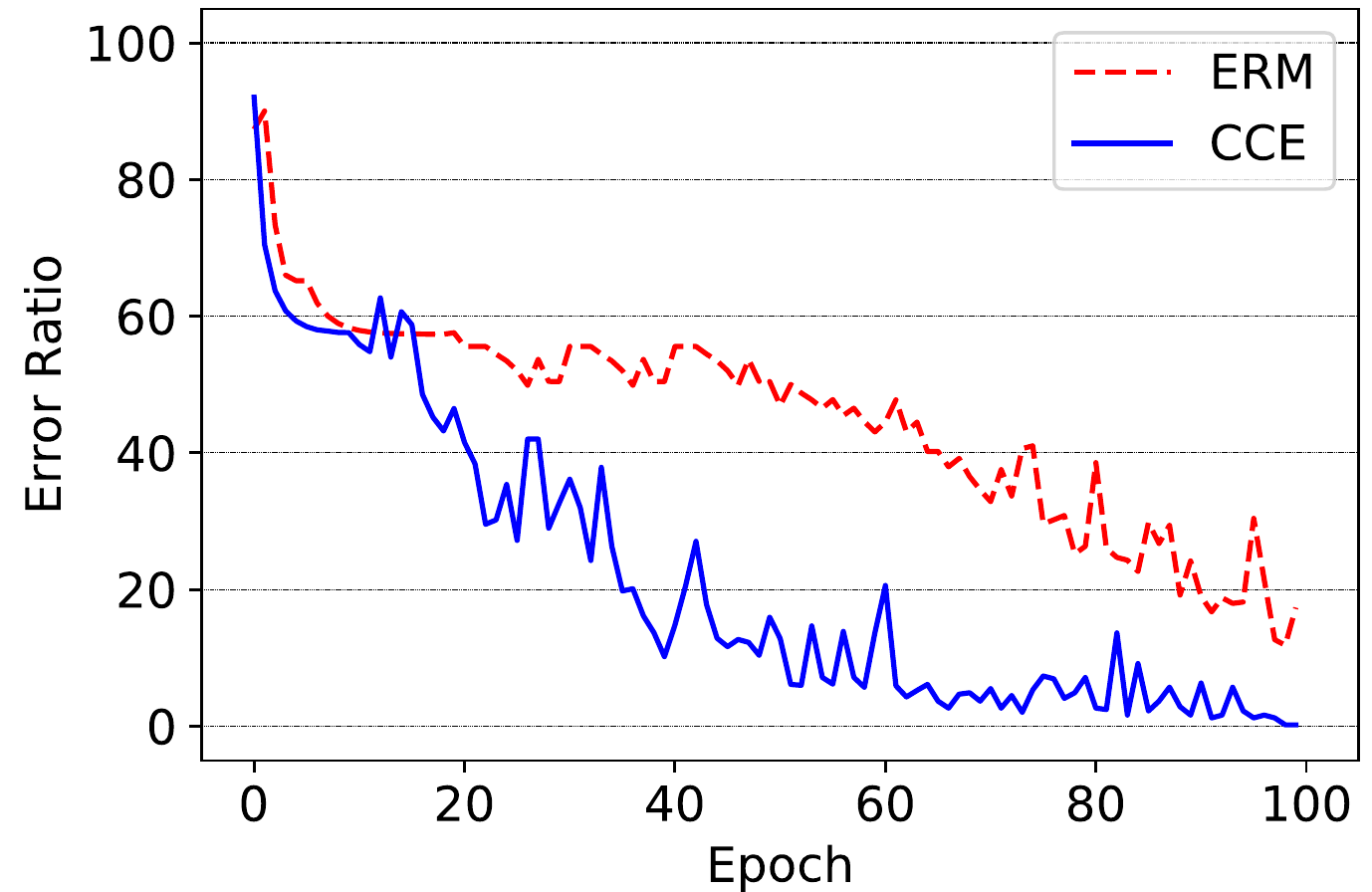}
    \caption{Test error (\%) of ResNet-101 on Road Marking. It can be seen that ResNet-101 with CCE converges much faster than one with ERM in terms of error rate.}
    \label{fig6}
    \vspace{-0.4cm}
    \end{figure}

    \textbf{Road Marking.} 
    It is consisted of colored 1,443 examples on road markings such as ``35,'' ``40,'' ``forward,'' and ``stop'' \cite{wu2012practical}. The number of classes in this dataset is 27. 
    All examples were taken on clear and sunny days. 
    This work splits the data into training set and test set at a ratio of 8:2.
    It originally has a long-tailed distribution as shown in Fig. \ref{fig2} (e).
    The distribution of classes on the test set is not balanced in the Road Marking. 
    For image classification experiments, this work crops out the backgrounds in all images. 
    Only examples of top 9 classes are used in experiments for fair comparison with the other state-of-the-art results. 
    The learning rate is set to 1e-2 and dropped by a factor of 0.5 at 40 and 80 epochs during training.
    As shown in Table \ref{road},  \ref{table5}, the proposed loss shows powerful results on Road Marking. 
    Especially in ResNet-101, the proposed loss achieves significant performance improvement of 11.02\% in terms of accuracy, compared to the ERM.
    Note that CCE also encourages the model converges faster than other methods such as ERM in terms of error ratio, as in Figure \ref{fig6}.
    
    \begin{table}[ht]
    \footnotesize
    \centering
    \caption{Classification test accuracy (\%) on Road Marking.}
        \begin{tabular}{@{}cllll@{}}
            \toprule
            Model & ERM & FL & COT & \textbf{CCE} \\ \midrule
            ResNet-50 & 98.78 & 98.18 & 98.78 & \textbf{99.18} \\
            ResNet-101 & 88.16 & 88.78 & 98.78 & \textbf{99.18} \\
            SqueezeNet & 94.69 & 95.59 & \textbf{96.53} & \textbf{96.53} \\
            EfficientNet\_b0 & 98.78 & 98.78 & 98.78 & \textbf{99.80} \\
            EfficientNet\_b1 & 99.18 & 99.18 & 99.18 & \textbf{99.39} \\
            EfficientNet\_b7 & 99.59 & 99.18 & \textbf{99.80} & \textbf{99.80} \\
            \bottomrule
        \label{road}
        \end{tabular}
        \vspace{-0.1cm}
    \end{table}
    
    \begin{table}[ht]
    \footnotesize
    \centering
    \caption{Comparison on classification test accuracy (\%) of the proposed method with other state-of-the-art methods on Road Marking.}
    \begin{tabular}{@{}ll@{}}
    \toprule
    Method                                                   &   \\ \midrule
    Lee \textit{et al.} (AlexNet + ERM) \cite{lee2019unconstrained}                         & 94.70          \\
    Lee \textit{et al.} (GAN + Augmentation) \cite{lee2019unconstrained} & 98.80          \\
    Bailo \textit{et al.} (PCANet + Logistic Regression) \cite{bailo2017robust}    & 98.90          \\
    Balio \textit{et al.} (PCANet + SVM) \cite{bailo2017robust}                    & 99.10          \\
    Ahmad \textit{et al.} (LeNet$_{96}$ CP$_{2}$) \cite{ahmad2017symbolic} & 99.05 \\
    \textbf{EfficientNet\_b0 + CCE}                                           & \textbf{99.80}          \\ \bottomrule
    \label{table5}
    \end{tabular}
    \vspace{-0.5cm}
    \end{table}

    \textbf{Street View House Numbers (SVHN).}
    It is consisted of images extracted from Google Street View \cite{netzer2011reading}.
    This work splits the data into training set with 73,257 digits and test set with 26,032 digits.
    This data originally has a long-tailed distribution as depicted in Fig. \ref{fig2} (j).
    Each pixel value on the whole images is then normalized into [-1, 1] for experiments.
    The results in table \ref{svhn} show that the proposed method outperforms the other methods in all experiments on SVHN.
    
    \begin{table}[ht]
    \footnotesize
    \centering
    \caption{Classification test accuracy (\%) on SVHN.}
        \begin{tabular}{@{}cllll@{}}
            \toprule
            Model & ERM & FL & COT & \textbf{CCE} \\ \midrule
            ResNet-50 & 94.78 & 94.76 & 94.97 & \textbf{95.15} \\
            ResNet-101 & 94.78 & 94.62 & 95.01 & \textbf{95.13} \\
            ResNeXt-50 & 93.08 & 93.47 & \textbf{93.76} & 93.60 \\
            Densenet-121 & 91.78 & 91.78 & 92.65 & \textbf{93.68} \\
            EfficientNet\_b0 & 92.45 & 92.98 & \textbf{93.01} & \textbf{93.01} \\
            EfficientNet\_b7 & 93.08 & 93.62 & 94.62 & \textbf{94.88} \\
            \bottomrule
        \label{svhn}
        \end{tabular}
        \vspace{-0.1cm}
    \end{table}

\subsection{Parameter ($\gamma$) Study}
    \begin{table}[ht]
    \footnotesize
    \centering
    \caption{Parameter ($\gamma$) study on variants of CIFAR-10 with ResNet-34.}
        \begin{tabular}{@{}cccccc@{}}
        \toprule
        \multirow{3}{*}{Parameter} & \multicolumn{5}{c}{CIFAR-10}                                                           \\ \cmidrule(l){2-6} 
                                   & \multirow{2}{*}{Original} & \multicolumn{2}{c}{Long-tailed} & \multicolumn{2}{c}{Step} \\ \cmidrule(l){3-6} 
                                   & & 10 & 100 & 10 & 100         \\ \midrule
        $\gamma=-1.0$ & \textbf{95.39} & \textbf{88.37} & \textbf{71.98} & \textbf{86.85} & 68.58  \\
        $\gamma=-2.0$ & 95.18 & 88.06  & 71.20 & 86.43 & \textbf{69.20}  \\
        $\gamma=-3.0$ & 95.09 & 88.12  & 71.14 & 86.48 & 67.21  \\
        $\gamma=-4.0$ & 95.21 & 87.66  & 70.54 & 86.31 & 65.23  \\
        $\gamma=-5.0$ & 95.20 & 87.57  & 70.19 & 86.18 & 63.80  \\
        $\gamma=-10.0$ & 94.81 & 87.00  & 69.97 & 85.86 & 63.54  \\
        $\gamma=-20.0$ & 94.32 & 86.01  & 68.54 & 84.97 & 62.53  \\
        $\gamma=-30.0$ & 93.30 & 85.00  & 61.93 & 82.33 & 55.78  \\
        $\gamma=-40.0$ & 92.13 & 84.01  & 61.29 & 80.30 & 56.04  \\
        $\gamma=-50.0$ & 91.96 & 83.16  & 51.08 & 76.34 & 54.21  \\ 
        \bottomrule
        \label{ablation}
        \end{tabular}
        \vspace{-0.5cm}
    \end{table}

    For an parameter study, this work reports the classification accuracy (bACC) on variants of CIFAR-10 by varying $\gamma$ on CCE. 
    As shown in Table \ref{ablation}, it can be seen that the smaller the $|\gamma|$ value is, the better the classification accuracy is.
    In other words, performance degradation occurs in terms of classification accuracy as the $|\gamma|$ value increases.
    Especially, the case when $\gamma$ is equal to $-1$ yields the best performance on almost all variants of CIFAR-10.
    Based on these observations, this work sets $\gamma$ to be $-1$ for all experiments.


\subsection{Weakness Analysis}
\begin{figure}[htbp]
\centering\includegraphics[width=7.7cm]{ 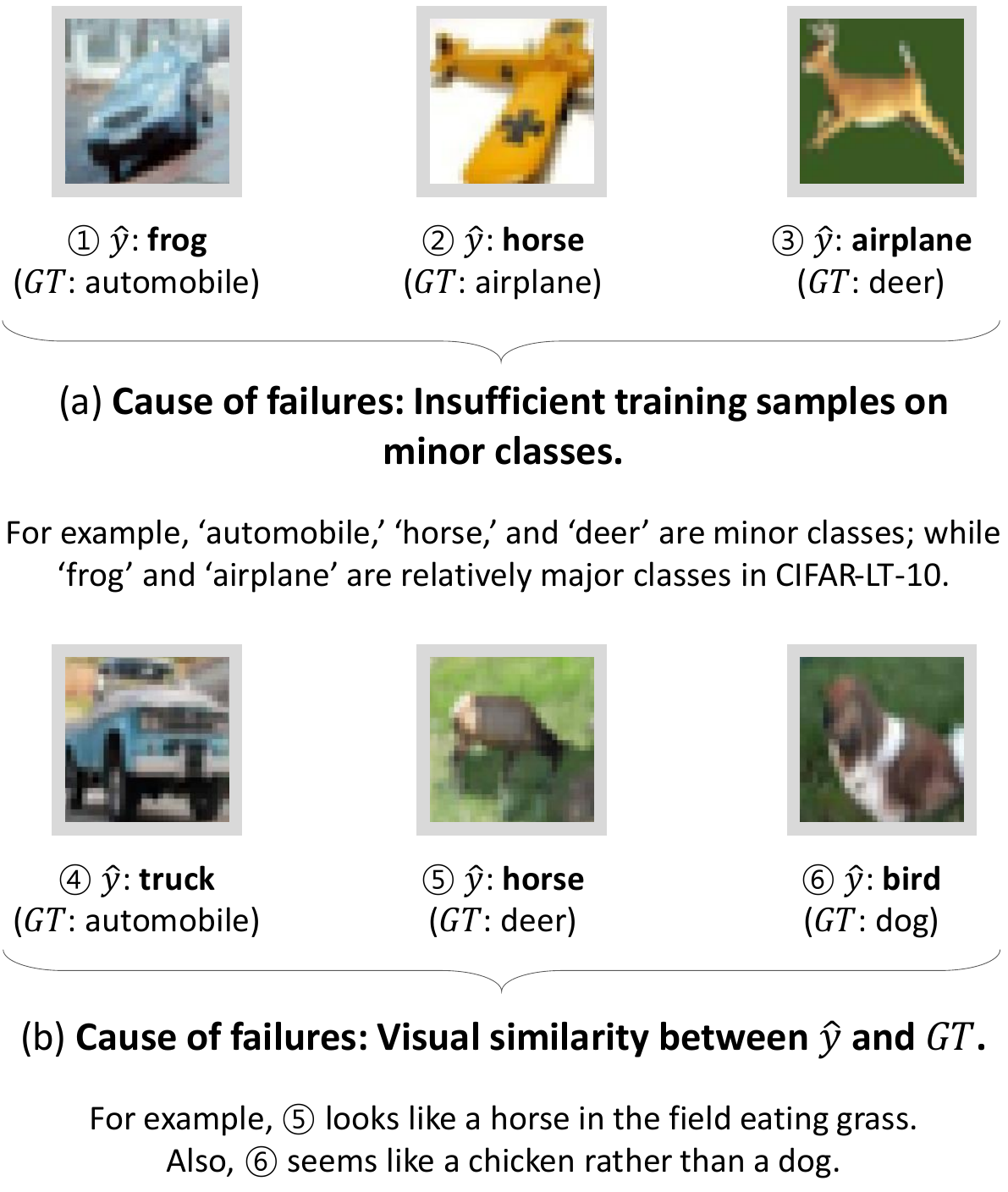}
\caption{Error study on imbalanced variants of CIFAR-LT-10. Let $\hat{y}$ and $GT$ denote, respectively, predicted class and ground truth class. The above images are examples misclassified by the best models, despite adopting the proposed loss function.}
\label{fig7}
\vspace{-0.5cm}
\end{figure}

    This subsection contains an error study on imbalanced variants of CIFAR-LT-10. 
    For these datasets, the proposed CCE almost enhances the classification performance in terms of prediction accuracy, compared to the other approaches. 
    However, there still exist incorrectly classified samples, despite adopting the proposed loss function (see Figure \ref{fig7}). 
    Although case (a) in Figure \ref{fig7} is hard to solve due to insufficient samples, this work argues that case (b) should be dealt with in further studies.
    The future work will address this problem by disambiguating visually similar categories.

\vspace{-0.0cm}
\section{Conclusion}\label{sec:4}
In this paper, we proposed a novel loss function, namely complement cross entropy (CCE) for imbalanced classification. 
This work proved that suppressing probabilities on incorrect classes helps the deep learning models to learn discriminative information. 
Especially with the proposed method, samples on minority classes are able to get better training opportunities by neutralizing highest softmax scores on wrong classes.
It also prevents overfitting to samples on majority classes or performance degradation in class-imbalanced datasets.
The proposed loss has shown powerful results on various image classification tasks.
In the future, this research can be extended by (i) mitigating the issue of visually confusing categories for further improvements and (ii) conducting additional experiments with considering various existing tricks for imbalanced classification.

\vspace{-0.1cm}
\section*{Acknowledgments}
\small
This work was partly supported by the Institute of Information $\&$ Communications Technology Planning $\&$ Evaluation (IITP) grant funded by the Korea government (MSIT) (No. 2014-3-00077), the National Research Foundation of Korea (NRF) grant funded by the Korea government (MSIT) (No. 2019R1A2C2087489), and the Korea Creative Content Agency (KOCCA) grant funded by the Korea government (MCST) (No. R2020070004).
Besides, the authors also gratefully thank the GIST Institute for AI (GIAI) for the support of the GPUs used in this work and Hyeongjun Yoo (jhdf1234@gm.gist.ac.kr) for reviewing this paper carefully.


\bibliographystyle{ieeetr}
\bibliography{refs}

%

\bio{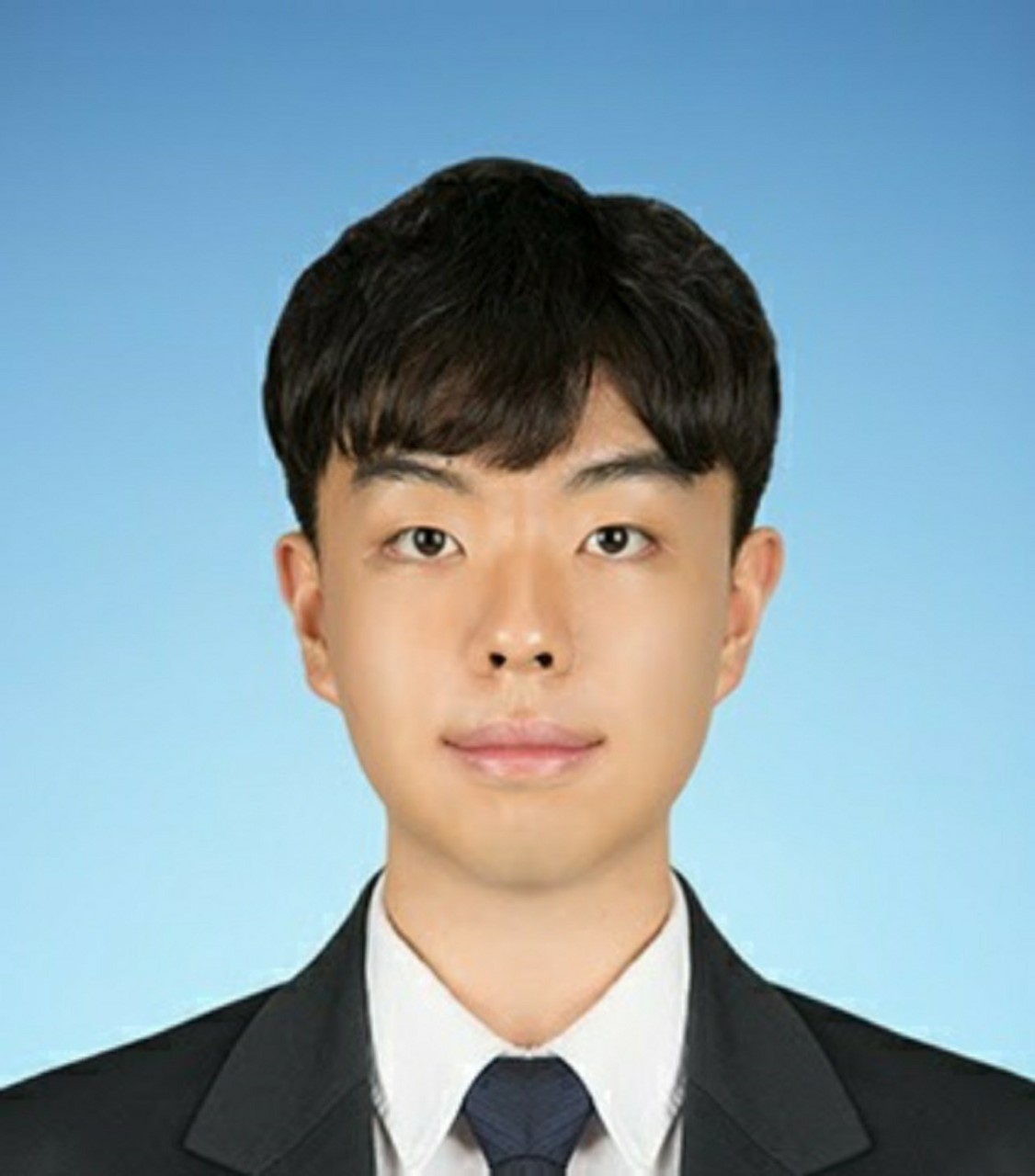}
\textbf{Yechan Kim} received the B.S. degree in College of Education from Jeju National University, South Korea, in 2020, and the M.S. degree in the School of Electrical Engineering and Computer Science from Gwangju Institute of Science and Technology (GIST), South Korea, in 2021.
He is currently pursuing the Ph.D. degree in artificial intelligence from Gwangju Institute of Science and Technology (GIST), South Korea. 
His current research interests include artificial intelligence, machine learning and computer vision.
\endbio

\bio{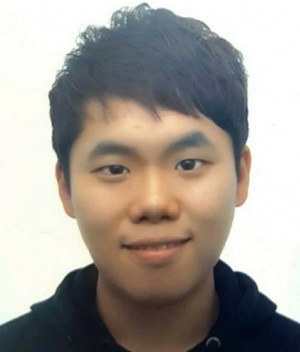}
\textbf{Younkwan Lee} received the B.S. degree in computer science from Korea Aerospace University, Gyeonggi, South Korea, in 2016. He is currently pursuing the Ph.D. degree in the School of Electrical Engineering and Computer Science, Gwangju Institute of Science and Technology (GIST), Gwangju, South Korea. His current research interests include computer vision, machine learning and deep learning.
\endbio

\bio{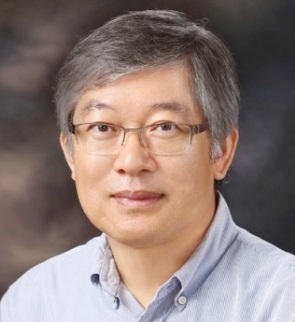}
\textbf{Moongu Jeon} received the B.S. degree in architectural engineering from Korea University, Seoul, South Korea, in 1988, and the M.S. and Ph.D. degrees in computer science and scientific computation from the University of Minnesota, Minneapolis, MN, USA, in 1999 and 2001, respectively. He was involved in optimal control problems with the University of California at Santa Barbara, Santa Barbara, CA, USA, from 2001 to 2003, and then moved to the National Research Council of Canada, where he was involved in the sparse representation of high-dimensional data and the image processing, until July 2005. In 2005, he joined the Gwangju Institute of Science and Technology, Gwangju, South Korea, where he is currently a Full Professor with the School of Electrical Engineering and Computer Science. His current research interests include machine learning, computer vision, and artificial intelligence.
\endbio

\end{document}